# Sparse Representation Based Augmented Multinomial Logistic Extreme Learning Machine with Weighted Composite Features for Spectral-Spatial Classification of Hyperspectral Images

Faxian Cao[1], Zhijing Yang[1*], Jinchang Ren[2], Wing-Kuen Ling[1]

[1] School of Information Engineering, Guangdong University of Technology, Guangzhou, 510006, China;

[2] Department of Electronic and Electrical Engineering, University of Strathclyde, Glasgow, G1 1XW, UK

*Email: yzhj@gdut.edu.cn

*Abstract*— Although extreme learning machine (ELM) has been successfully applied to a number of pattern recognition problems, it fails to provide sufficient good results in hyperspectral image (HSI) classification due to two main drawbacks. The first is due to the random weights and bias of ELM, which may lead to ill-posed problems. The second is the lack of spatial information for classification. To tackle these two problems, in this paper, we propose a new framework for ELM based spectral-spatial classification of HSI, where probabilistic modelling with sparse representation and weighted composite features (WCF) are employed respectively to derive the optimized output weights and extract spatial features. First, the ELM is represented as a concave logarithmic likelihood function under statistical modelling using the maximum a posteriori (MAP). Second, the sparse representation is applied to the Laplacian prior to efficiently determine a logarithmic posterior with a unique maximum in order to solve the ill-posed problem of ELM. The variable splitting and the augmented Lagrangian are subsequently used to further reduce the computation complexity of the proposed algorithm and it has been proven a more efficient method for speed improvement. Third, the spatial information is extracted using the weighted composite features (WCFs) to construct the spectral-spatial classification framework. In addition, the lower bound of the proposed method is derived by a rigorous mathematical proof. Experimental results on two publicly available HSI data sets demonstrate that the proposed methodology outperforms ELM and a number of state-of-the-art approaches.

## I. INTRODUCTION

With rich spectral and spatial information contained in a three-dimensional hypercube, hyperspectral images (HSI) provides a unique way for characterizing objects in geographical scenes, especially remote sensing images [1]. However, classification of high dimensional data such as HSI is still challenging, particularly due to the unfavorable ratio between the limited number of training samples and large number of spectral bands, i.e. the Hughes phenomenon [2]-[4]. To tackle this problem, a number of feature extraction and data classification approaches have been proposed [12]. These include the singular spectrum analysis (SSA) [5]-[8], segmented auto-encoders [11], principal component analysis (PCA) and its variations [13]-[14], spectral-spatial classification using multiple kernels and active learning [15][16], where support vector machines (SVM) [9][10] are widely used for data classification. Although these approaches have produced quite good results, their performance can be further improved by addressing two main difficulties: (1) Inaccurate classification under a large number of spectral bands yet limited training samples. (2) Relatively low efficiency for processing high dimensional HSI data.

As a single forward layer neural network, the extreme learning machine (ELM) is a fast and effective machine learning method and has received a wide attention due to its good performance [17]-[19]. The ELM needs not tune the hidden layer parameters once the number of hidden layer nodes is determined. In ELM, the weight and bias vectors between the input layer and the hidden layer are randomly generated, which are independent of the training samples and the specific applications [1]. ELM have achieved good performance in many applications [20-23] due to its good performance. Also, ELM has been widely applied to HSI classification. For example, In [24],[25], bilateral filtering and extended morphological profiles were used for feature extraction respectively. And ELM was used for classification. In [26],[27], [28], superpixel, watershed and Gabor filter were used for feature extraction and ELM was used for classification. Although these methods ELM-based have achieved good performance, they just combine ELM and other feature extraction method. These ELM-based method ignore one important problem in ELM that the randomly generated input weights and bias of ELM may cause ill-posed problems. Based on this perspective, we propose augmented sparse multinomial logistic ELM (ASMLELM) for HSIs classification. Based on proposed ASMLELM, we propose weighted composite features (WCFs) for extracting rich spatial information for ASMLELM. Therefore, a novel framework ASMLELM-WCFs have been proposed for spectral-spatial classification of HSI, and the main contributions can be highlighted as follows.

First, the ELM is represented by a maximum a posteriori (MAP) based probabilistic model, which is further represented by a concave logarithmic likelihood function (LLF). As the LLF value can be arbitrarily large if the training data is separable, a prior/regularized term on the LLF is critical [29], for which the sparse representation is employed for representing the ELM. In order to settle the ill-posed problem in ELM, Laplacian prior/regularized term is applied to improve the sparsity of the learnt weights of the proposed ELM classifier. The Laplacian function has the heaviest tailed density derived from its logarithmic concave, where its sparsity promoting nature has been theoretically well justified [29]-[32]. This can guarantee the logarithmic posterior to have a unique maximum when it is combined with a concave logarithmic likelihood [29]. As the ELM is represented by a probabilistic model under a maximum a posteriori (MAP), it can enjoy the advantage of existing a unique maximum. Therefore, the ill-posed problem in ELM can be addressed by above operation namely spare multinomial logistic ELM (SMLELM). Recently, variable splitting and augmented Lagrangian [33] have been proposed for speed improvement. We adopt the idea variable splitting and augmented Lagrangian for proposed SMELM to construct proposed augmented SMLELM (ASMLELM), which transform SMLELM into a new form using variable splitting and augmented Lagrangian for speed improvement.

Second, by combining composite kernels (CK) [34] and weighted mean filters (WMFs) [35], the weighted composite features (WCFs) are utilized to extract spatial features and further improve the classification accuracy. Accordingly, three improved spectral-spatial classifiers are derived, which include the ELM, the nonlinear ELM (NLELM) and the kernel ELM (KELM) based classifiers.

Third, the generalization bounds of the proposed method is derived in a similar way to determine the margin bounds of the sparse multinomial logistic regression (SMLR) [29][36][37]. These bounds can provide a theoretical insight of and justification for our proposed methods.

The rest of this paper is organized as follows. In Section II, the background of the ELM is introduced. The proposed method is detailed in Section III. Section IV reports the experimental results in benchmarking with a few state-of-the-art approaches. Finally, some conclusions are drawn in Section V.

## II. THE EXTREME LEARNING MACHINE (ELM)

### A. Basic Concepts of ELM

The ELM is a generalized single layer feedforward neural network (SLFNs) [1], [17]. The weight vector and the bias between input layer and hidden layer are randomly initialized, though the final values will be determined by the learning process. Once the initial values for the weight/bias vectors are assigned, the hidden layer output matrix remains unchanged in the learning process [1].

Let $X \equiv (x_1, x_2, \ldots, x_N) \in R^{d \times N}$ be the training data of a HIS, which has $N$ pixels and each pixel has a $d$-dimensional feature. Let $Y = (y_1, y_2, \ldots, y_N) \in R^{M \times N}$ be a matrix representing the class label of the training samples, where M is the number of class in datasets. Given a pixel label $y_i$, if it belongs to the $k$-th class, we have

$$y_{i,j} = \begin{cases} 1, & j = k, \\ 0, & otherwise. \end{cases}$$

The model of a single hidden layer of the neural network with $L$ hidden neurons and the activation function $H(x)$ can be expressed as follows:

$$\sum_{j=1}^{L} \beta_j H(w_j^T x_i + b_j) = y_i, \text{ i=1,2,…,N} \tag{1}$$

where $\beta_j$ represents the weight vector between the hidden layer and the output layer; $w_j$ and $b_j$ are the weight vector and bias from the inputs to the hidden layer, respectively; $H(w_j^T x_i + b_j)$ represents the output of the $j$-th hidden neuron with respect to the input sample $x_i$. Obviously, equation (1) can be further expressed in the following matrix form:

$$H^T \beta = Y^T \tag{2}$$

where $\beta = [\beta_1 \cdots \beta_M]_{L \times M}$, $H = [H(x_1) \cdots H(x_n)]_{L \times N}$, and $H(x_i) = [H_1(x_i) \cdots H_L(x_i)]_{L \times 1}^T$. $H$ is the hidden layer output matrix, and $\beta$ is the output weight matrix between hidden layer and output layer.

From (2), $\beta$ can be simply obtained below, where † is the Moore Penrose generalized inverse of a matrix [17].

$$\beta = (H^T)^{\dagger} Y^T \tag{3}$$

### B. Constrained Optimization of the ELM

The constrained optimization of the ELM aims to achieve not only the smallest training error but also the smallest output weights [19]:

$$\min \parallel H^T \beta - Y^T \parallel^2 \text{ and } \parallel \beta \parallel^2 \tag{4}$$

According to the Bartlett's neural network generalization theory [38], the smaller weights will result in a smaller training error of the feedforward neural networks. According to this optimization theory, Eq. (4) can be rewritten as:

$$\min_{\beta, \xi_i} L_{ELM} = \frac{1}{2} \parallel \beta \parallel_F^2 + C \frac{1}{2} \sum_{i=1}^{N} \parallel \xi_i \parallel_2^2, \quad \text{s.t.} H^T(x_i)\beta = y_i^T - \xi_i^T, \ i=1,\ldots,N \tag{5}$$

where $\xi_i$ is the training error for the training sample $x_i$, $C$ is the regularization parameter.

Based on the Karush-Kuhn-Tucker (KKT) theorem [39], training the ELM is equivalent to solve the following dual optimization problem:

$$\min_{(\beta,\alpha,\xi_i)} L_{ELM} = \frac{1}{2} \parallel \beta \parallel_F^2 + C \frac{1}{2} \sum_{i=1}^{N} \parallel \xi_i \parallel_2^2 - \sum_{i=1}^{N} \sum_{j=1}^{M} \alpha_{i,j} \left( H^T(x_i)\beta_j - y_{i,j} + \xi_{i,j} \right) \tag{6}$$

where $\beta_j$ is the column vector of the matrix $\beta$, and $\alpha_{i,j}$ is the Lagrange multiplier.

From the KKT theorem, we can further derive

$$\frac{\partial L_{ELM}}{\partial \beta_j} = 0 \rightarrow \beta = H * \alpha \tag{7}$$

$$\frac{\partial L_{ELM}}{\partial \varepsilon_i} = 0 \rightarrow \alpha_i = C\varepsilon_i, \quad i = 1, \dots, N \tag{8}$$

$$\frac{\partial L_{ELM}}{\partial \alpha_i} = 0 \rightarrow H^T(x_i)\beta = y_i^T - \xi_i^T \quad i = 1, \dots, N \tag{9}$$

where $\alpha_i = [\alpha_{i,1}, \alpha_{i,2}, \dots, \alpha_{i,M}]^T$ and $\alpha = [\alpha_1, \alpha_2, \dots, \alpha_N]^T$.

Then, it can be shown that the output weight $\beta$ is:

$$\beta = H \left( \frac{I}{C} + H^T H \right)^{-1} Y^T \tag{10}$$

The activation functions of the neurons in the hidden layer are unknown, and any kernel satisfying the Mercer's conditions can be used:

$$\begin{cases} \mathbf{\Omega}_{KELM} = H^T * H \\ \Omega_{KELM}(x_i, x_j): \mathrm{h}(x_i)^T \mathrm{h}(x_j) = K(x_i, x_j) \end{cases} \tag{11}$$

In fact, the Gaussian kernel is one of the good choices:

$$K_{ELM}(x_i, x_j) = \exp(-\frac{\|x_i - x_j\|^2}{2*\sigma_{ELM}}) \tag{12}$$

Based on the above analysis, two well-known constrained optimization methods of ELM had been proposed. One is to define $\beta$ in (10) without a kernel, namely nonlinear ELM (NLELM). The other one is to use the kernel function for the kernel ELM (KELM) as given below.

$$\beta_{NLELM} = H(\frac{I}{C} + H^T H)^{-1} Y^T \tag{13}$$

$$\beta_{KELM} = (\frac{I}{C} + K(x_i, x_j))^{-1} Y^T \tag{14}$$

### III. The Proposed Framework

#### A. Augmented Sparse Multinomial Logistic Extreme Learning Machine (ASMLELM)

The goal of a supervised learning algorithm is to design a classifier based on a set of N training samples that is capable of distinguishing M classes on the basis of an input vector of length d [29]. Under the multinomial logistic regression model [40], $\beta$ in equations (3), (13) and (14) can be transformed to a new form via a probability model. If training sample $x_i$ belongs to the j-th class, the probability model can be represented by the following equation:

$$P(y_{i,j} = 1 | H(x_i), \beta) = \frac{exp(\beta_j^T H(x_i))}{\sum_{j=1}^{M} exp(\beta_j^T H(x_i))} \tag{15}$$

In (3), (13), (14) and (15), $\beta$ may not be optimal due to the ill-posed problem. Therefore, it is important to find the optimal $\beta$ to obtain high classification accuracy. In order to find the optimal parameter $\beta$ for the ELM, $\beta$ will be estimated again after presenting the ELM by a probabilistic model. To this end, the maximum likelihood (ML) estimation is introduced to the ELM. Let $\beta = [\beta_1; \beta_2; \cdots; \beta_M]_{(L \times M) \times 1}$ be a column vector with $L \times M$ elements. A simple maximization of the logarithmic likelihood can be expressed as follows:

$$\max_{\beta} L(\beta) = \sum_{i=1}^{N} (\sum_{j=1}^{M} y_{i,j}\beta_j^T H(x_i) - log \sum_{j=1}^{M} exp(\beta_j^T H(x_i))) \tag{16}$$

In order to maximize $L(\beta)$, consider the second order Taylor series of $L(\beta)$ evaluated at $\beta'$:

$$L(\beta) - L(\beta') = (\beta - \beta')^T \nabla L(\beta') + \frac{1}{2}(\beta - \beta')^T \nabla^2 L(\beta' + \rho(\beta - \beta'))(\beta - \beta')$$

$$\geq (\beta - \beta')^T \nabla L(\beta') + \frac{1}{2}(\beta - \beta')^T B(\beta - \beta') \tag{17}$$

where $\rho \in (0,1)$ and

$$B \equiv -\frac{1}{2}[\mathbf{I} - \frac{\mathbf{11}^T}{M}] \otimes HH^T \tag{18}$$

where $\mathbf{I} \in R^{M \times M}$ is an identity matrix, $\mathbf{1} = [1, 1, \dots, 1]^T$ and $\otimes$ is the kronecker matrix product. This result has been proved in [40], [41]. Then, the ML estimation can be expressed as follows:

$$\hat{\beta} = arg \max_{\beta} \beta^T(\nabla L(\beta') - B\beta') + \frac{1}{2}\beta^T B\beta \tag{19}$$

Hence, $\beta$ at the (t+1)-th iteration can be expressed by a simple update equation:

$$\hat{\beta}^{t+1} = B^{-1}(B\hat{\beta}^t - \nabla L(\hat{\beta}^t)) \tag{20}$$

From (20), it can be seen that it is very similar to an iteratively reweighted least squares (IRLS) algorithm [42]. However, the Hessian matrix in the IRLS algorithm is replaced by matrix B. Since $B^{-1}$ can be precomputed, it is a big advantage of the proposed approach. Compared to the IRLS algorithm, whose Hessian matrix must be inverted at each iteration [29], [43], our proposed approach is better than the IRLS algorithm.

However, the concave LLF value can be arbitrarily large if the training data is separable. From [29], it is known that a prior on the logarithmic likelihood is crucial. In order to address the ill-posed problem in ELM, the prior/regularized term is adopted on $\beta$. Here, the Laplacian prior is used:

$$L1(\beta) = L(\beta) - L(\beta') + logp(\beta) \tag{21}$$

where

$$p(\beta) \propto exp(-\lambda \parallel \beta \parallel_1) \tag{22}$$

and $\parallel \beta \parallel_1 = \sum_l |\beta_l|$ denotes the $l_1$ norm and $|\beta_l| = \sqrt{\beta_l^2}$.

Consider the following inequality:

$$h + u \geq 2\sqrt{h}\sqrt{u} \xrightarrow{yields} \sqrt{u} \leq \frac{1}{2}(\frac{u}{\sqrt{h}} + \sqrt{h}) \tag{23}$$

where h>0 and u>0.

For any $\beta'$, we have

$$-\lambda \parallel \beta \parallel_1 \geq -\frac{1}{2}\lambda(\sum_l \frac{\beta_l^2}{|\beta_l'|} + \sum_l |\beta_l'|) \tag{24}$$

Therefore,

$$\beta^T(\nabla L1(\beta') - B\beta') + \frac{1}{2}\beta^T(B - \lambda\Lambda)\beta \tag{25}$$

can be maximized. Here,

$$\Lambda = diag\{|\beta_{11}|^{-1}, ..., |\beta_{LM}|^{-1}\} \tag{26}$$

Finally, (20) can be expressed by the following equation:

$$\hat{\beta}^{t+1} = (B - \lambda\Lambda^t)^{-1}(B\hat{\beta}^t - \nabla L(\hat{\beta}^t)) \tag{27}$$

From the above, it can be seen that the Laplacian prior/regularized term is applied to $\beta$ with $\lambda$ acting as a regularization parameter. The Laplacian prior imposed on the sparse multinomial logistic ELM (SMLELM) controls the complexity of the SMLELM classifier and improves the generalization capacity of the SMLELM since $p(\beta)$ in (22) forces many components of $\beta$ to be zero.

Since the term L($\beta$) in (16) is not quadratic and $p(\beta)$ in (22) is nonsmooth, finding the solution of the optimization problem defined as (25) is very difficult. Recently, a novel method called the majorization minimization [43] has been proposed to decompose this kind of problems [29], [44]-[47]. However, the computation complexity of this algorithm is very large. In [48], the logistic regression via a variable splitting and an augmented Lagrangian (LORSAL) have been used for speed improvement. Moreover, variable splitting and augmented Lagrangian have been used for speed improvement in some work [33] [16] [48]. This algorithm have been proven that can greatly reduce the computation complexity [16], [45], [49]. As the variable splitting and augmented Lagrangian approach shows a good performance in the corresponding works, then we can use this approach to reduce the time-consuming of proposed SMLELM, which transform proposed SMLELM into a new form.

Variable splitting is a very simple procedure which consist in creating a new variable[50]. Then the problem defined in (21) is equivalent to:

$$(\hat{\beta}, \hat{v}) = arg \min_{\beta,v} -L(\beta) + \lambda \parallel v \parallel_1 \text{ s.t. } \beta = v \tag{28}$$

This optimization problem can be solved via applying the direction method of multipliers [51] (see also [52] and the references therein).This neural network is called the augmented SMLELM(ASMLELM).

Apply the augmented Lagrangian [50] to solve the equation (28). Then, the solution of (28) at (t+1)-th iteration can be rewritten as follows:

$$\hat{\beta}^{t+1} = arg \min_{\beta} -L(\beta) + \frac{\gamma}{2} \parallel \beta - v^t - b^t \parallel^2 \tag{29}$$

$$\hat{v}^{t+1} = arg \, arg \min_{v} \lambda \parallel v \parallel_1 + \frac{\gamma}{2} \parallel \beta^t - v - b^t \parallel^2 \tag{30}$$

and

$$b^{t+1} = b^t - \beta^{t+1} + v^{t+1} \tag{31}$$

where $\gamma \geqq 0$ is the weight of the augmented SMLELM (ASMLELM). For any $\gamma \geqq 0$, the sequence $\hat{\beta}^t$ converges to a minimizer [45], [48] [50]. For the convenience, $\gamma = 10\lambda$ is set. The solution of the problem defined in (28) is the simple soft-threshold rule [51], [45]. It can be expressed as:

$$v^{t+1} = \max(0, abs(e) - \frac{1}{10}\} \tag{32}$$

where

$$e = \beta^{t+1} - b^t \qquad (33)$$

After designing the ASMLELM framework which can be applied to the ELM, NLELM and KELM, three new spectral algorithms for performing the HSI classification can be generated. They are named as the ASMLELM for the basic ELM (ASMLBELM), the ASMLELM for the NLELM (ASMLNLELM) and the ASMLELM for the KELM (ASMLKELM). The pseudo-docodes for these three methods are shown in Algorithm 1.

---

**Algorithm 1:The ASMLELM for basic the ELM, the NLELM and the KELM**

Input: The training sample pairs $\{x = (x_1,\ x_2,\ \dots, x_n\ )$ and $Y = (y_1, y_2, \dots, y_n)\}$ as well as the parameters $\lambda$, $b = 0$.

Training phase

    L: The number of nodes in a hidden layer.

    $H(\bullet)$: The sigmoid function.

    $\beta$: The output weight in the hidden layer.

1: Randomly generate the input weight $\{w_1, \dots w_L\}$ and the bias $\{b_1, \dots, b_L\}$.

2: For each training sample $\boldsymbol{x_i}$, calculate the hidden layer matrix

    $H(x_i) = [H_1(w_1 * \boldsymbol{x_i} + b_1), \dots, H_L(w_L * \boldsymbol{x_i} + b_L)]^T_{L \times 1}$.

3: Calculate the output weight

  (1) $\beta = (H^T)^{\dagger} Y^T$ for the ASMLBELM.

  (2) $\min\limits_{\beta, \xi_i} L_{ELM} = \frac{1}{2} \parallel \beta \parallel^2_F + C \frac{1}{2} \sum_{i=1}^{N} \parallel \xi_i \parallel^2_2$,

    s.t. $H^T(x_i)\beta = y_i^T - \xi_i^T$   i=1,...N,.

  Here, $\beta = H * \left(\frac{I}{C} + H^T H\right)^{-1} Y^T$ for the ASMLNLELM.

  (3) Let $\pi = \left(\frac{I}{C} + H^T H\right)^{-1} Y^T$ and the Gaussian kernel

  $K_{Train}(x_i, x_j) = \exp(-\frac{\parallel x_i - x_j \parallel^2}{2 * \sigma_{ELM}})$.

  Then, $\pi = \left(\frac{I}{C} + K_{Train}\right)^{-1} Y^T$ for the ASMLKELM.

4. Represent the ELM by a probability model

  (1) $P\left(y_{i,j} = 1 \middle| H(x_i), \beta\right) = \frac{exp(\beta_j^T H(x_i))}{\sum_{j=1}^{M} exp(\beta_j^T H(x_i))}$ for the ASMLBELM and the ASMLNLELM.

  (2) $P\left(y_{i,j} = 1 \middle| K_{train}(x_i), \pi\right) = \frac{exp(\pi_j^T K_{Train})}{\sum_{j=1}^{M} exp(\pi_j^T K_{Train})}$ for the ASMLKELM.

5. ASMLELM: The ML estimate based on the sparse representation with the Laplacian prior and the LORSAL algorithm.

5.1 $\hat{\beta} = arg \max\limits_{\beta} \beta^T (\nabla L(\beta') - B\beta') + \frac{1}{2} \beta^T (B - \lambda \Lambda^t) \beta$.

5.2 Set t=0.

5.3 Repeat.

5.4 $\hat{\beta}^{t+1} = \arg \min\limits_{\beta} -L(\beta) + \frac{10\lambda}{2} \parallel \beta - v^t - b^t \parallel^2$.

5.5 $\hat{v}^{t+1} = \arg \arg \min\limits_{v} \lambda \parallel v \parallel_1 + \frac{10\lambda}{2} \parallel \beta - v - b^t \parallel^2$.

5.6 $b^{t+1} = b^t - \beta^{t+1} + v^{t+1}$.

5.7 Increase $t$ to $t+1$; If the ASMLKELM is applied, replace $\beta$ by $\pi$.

5.8 Quit the algorithm until the stopping criterion is met.

    Prediction phase $\{X \equiv (x_1,\ x_2,\ \dots, x_N\ )$ and $Y = (y_1, y_2, \dots, y_N)\}$.

1: (1) Calculate the output layer matrix

    $H^*(\boldsymbol{x_i}) = [H_1(w_1 x_i + b_1),\ \dots,\ H_L(w_L x_i + b_L)]^T_{L \times 1}$   i=1,...,N for the ASMLBELM and the ASMLNLELM.

  (2) $K_{test} = H^{*T} H = \exp(-\frac{\parallel x_i - x_j \parallel^2}{2\sigma_{ELM}})$ for the ASMLKELM.

2: (1) $P\left(y_{i,j} = 1 \middle| H^*(x_i), \beta\right) = \frac{exp(\beta_k^T H^*(x_i))}{\sum_{k=1}^{M} exp(\beta_k^T H^*(x_i))}$ for the ASMLBELM and the AMSLNLELM.

  (2) $P\left(y_{i,j} = 1 \middle| K_{test}, \pi\right) = \frac{exp(\pi_k^T K_{test})}{\sum_{k=1}^{M} exp(\pi_k^T K_{test})}$ for the ASMLKELM.

---

*B. Weighted Composite Features Based ASMLELM (ASMLELM-WCFs)*

From the above, it can be seen that the ASMLELM can just use the spectral information of the HSI data for classification. A pixel and its spatial neighborhood pixel likely belong to the same class [1]. Therefore, the spatial information is very important. The spatial information will be adopted to the ASMLELM. Hence, the WCFs will be used to perform the spectral spatial classification for the proposed ASMLELM.

For a given pixel $x_i$, let the pixel coordinate of sample $x_i$ be (p, q), then the local pixel neighborhood centered at $x_i$ is $N(x_i) = \{x = (p,q)|p \in [p-a, p+a]; q \in [q-a; q+a]\}$, a=($w_{opt}$-1)/2 where $w_{opt}$ is the width of the neighborhood window. Let $x_i^w$ be the spectral feature of the training sample and $x_i^s$ be the information extracted from a local spatial neighborhood of the pixel $x_i^w$. It can be represented as $x_i^s = \sum_{x_k^w} x_k^w v_k / \sum_{x_k^w} v_k$ with the weight $v_k = \exp\{-z \parallel x_i - x_{ik} \parallel^2\}$ measuring the spectral distance between the central pixel and the neighboring pixels ($x_{ij} \in N(x_i)$). Following the setting in [35], we set $z = 0.2$ in this work.

Then, the output matrix of the hidden layer defined in (2) and (13) can be expressed as:

$$H = \mu H_w + (1-\mu)H_s \tag{34}$$

where

$$H_w = [H_w(x_1^w) \quad \cdots \quad H_w(x_n^w)]_{L \times N} \tag{35}$$

$$H_s = [H_s(x_1^s) \quad \cdots \quad H_s(x_n^s)]_{L \times N} \tag{36}$$

and μ is a combination coefficient balancing the spectral and spatial information.

For the KELM defined in (14), let $\beta$ be

$$\beta = (\frac{I}{C} + K)^{-1} Y^T \tag{37}$$

where

$$K = \mu K_{H_w} + (1-\mu)K_{H_s} \tag{38}$$

$$K_{H_w}(x_i^w, x_j^w) = \exp(-\frac{\parallel x_i^w - x_j^w \parallel^2}{2 * \sigma_w}) \tag{39}$$

and

$$K_{H_s}(x_i^s, x_j^s) = \exp(-\frac{\parallel x_i^s - x_j^s \parallel^2}{2 * \sigma_s}) \tag{40}$$

Here, $\sigma_w$ and $\sigma_s$ control the widths of the spectral and spatial Gaussian kernel. Now, three new methods for performing the spectral-spatial HSI classification can be proposed via the ASMLELM and WCFs. That is, the ASMLBELM-WCFs, the ASMLNLELM-WCFs and the ASMLKELM-WCFs. The details are summarized in Algorithm 2. Figure 1 show the flowchart of proposed ASMLELM-WCFs.

---

**Algorithm 2: ASMLELM with WCFs (ASMLELM-WCFs)**

Input: The spectral feature $\{x^w = (x_1^w, x_2^w, \ldots, x_n^w)$ and $Y = (y_1, y_2, \ldots, y_n)\}$, the spatial feature $\{x^s \equiv (x_1^s, x_2^s, \ldots, x_n^s)$ and $Y = (y_1, y_2, \ldots, y_n)\}$ as well as the parameters C, $\lambda$, b = 0.

Training phase:

    L: The number of nodes in a hidden layer.

    $H(\bullet)$: The sigmoid function.

    The output weight of the hidden layer $\beta$.

1: Randomly generate the input weight $\{w_1, \ldots w_L\}$ and the bias $\{b_1, \ldots, b_L\}$.

2: For any training sample $x_i$, calculate the hidden layer matrix
$H_w(x_i^w) = [H_1(w_1 * x_i^w + b_1), \ldots, H_L(w_L * x_i^w + b_L)]_{L \times 1}^T$ and $H_s(x_i^s) = [H_1(w_1 * x_i^s + b_1), \ldots, H_L(w_L * x_i^s + b_L)]_{L \times 1}^T$. Here,

    (1) $H = \mu H_w + (1-\mu)H_s$ for the ASMLBELM-WCFs.

    (2) $H = \sqrt{\mu}H_w + \sqrt{(1-\mu)}H_s$ for the ASMLNLELM-WCFs and the ASMLKELM-WCFs.

3: Calculate the output weight

    (1) $\beta = (H^T)^\dagger Y^T$ for the ASMLBELM-WCFs.

    (2) $\min_{\beta, \xi_i} L_{ELM} = \frac{1}{2} \parallel \beta \parallel_F^2 + C\frac{1}{2}\sum_{i=1}^N \parallel \xi_i \parallel_2^2$, s.t. $H^T(x_i)\beta = y_i^T - \xi_i^T$ i=1,...,N.

    Here, $\beta = H * (\frac{I}{C} + H^T H)^{-1} Y^T$ for the ASMLNLELM-WCFs.

    (3) $\pi = (\frac{I}{C} + K_{train})^{-1}Y^T$, $K_{train} = \mu H_w^T H_w + (1-\mu)H_s^T H_s = \mu K_{H_w} + (1-\mu)K_{H_s}$ for the ASMLKELM-WCFs.

4.Represent the ELM by a probability mode

$P(y_{i,j} = 1|H(x_i), \beta) = \frac{exp(\beta_j^T H(x_i))}{\sum_{j=1}^M exp(\beta_j^T H(x_i))}$ for the ASMLBELM-WCF and the ASMLNLELM-WCFs.

$P(y_{i,j} = 1|K_{train}(x_i), \pi) = \frac{exp(\pi_j^T C K_{train}(x_i))}{\sum_{j=1}^M exp(\pi_j^T K_{train}(x_i))}$ for the ASMLKELM-WCFs.

5. ASMLELM : The ML estimate based on the sparse representation with the Laplacian prior and the LORSAL

algorithm.

    5.1 $\hat{\beta} = arg \max_{\beta} \beta^T (\nabla L(\beta') - B\beta') + \frac{1}{2}\beta^T(B - \lambda\Lambda^t)\beta$.

    5.2 t:=0.

    5.3 Repeat.

    5.4 $\hat{\beta}^{t+1} = \arg \min_{\beta} -L(\beta) + \frac{10\lambda}{2} \parallel \beta - v^t - b^t \parallel^2$.

    5.5 $\hat{v}^{t+1} = \arg \arg \min_{v} \lambda \parallel v \parallel_1 + \frac{10\lambda}{2} \parallel \beta - v - b^t \parallel^2$.

    5.6 $b^{t+1} = b^t - \beta^{t+1} + v^{t+1}$.

    5.7 Quit the algorithm until the stopping criterion is met.

 Prediction phase:

The spectral feature $\{X^w = (x_1^w, x_2^w, \ldots, x_N^w)$ and $Y = (y_1, y_2, \ldots, y_N)\}$ as well as the spatial feature $\{X^s = (x_1^s, x_2^s, \ldots, x_N^s)$ and $Y = (y_1, y_2, \ldots, y_N)\}$.

1: Calculate the output layer matrix

$H_w^*(x_i^w) = [H_1(w_1 * x_1^w + b_1), \ldots, H_L(w_L * x_i^w + b_L)]_{L\times 1}^T$

and

$H_s^*(x_i^s) = [H_1(w_1 * x_i^s + b_1), \ldots, H_L(w_L * x_i^s + b_L)]_{L\times 1}^T$. Here,

    (1) $H^* = \mu H_w^* + (1-\mu)H_s^*$ for the ASMLBELM-WCFs.

    (2) $H^* = \sqrt{\mu}H_w^* + \sqrt{(1-\mu)}H_s^*$ for the ASMLNLELM-WCFs.

    (3) $H^* = \sqrt{\mu}H_w^* + \sqrt{(1-\mu)}H_s^*$, $K_{test} = \mu H_w^{*T}H_w^* + (1-\mu)H_s^{*T}H_s^* = \mu K_{H_w} + (1-\mu)K_{H_s}$ for ASMLKELM-WCFs.

2:  (1) $P(y_{i,j} = 1 | H^*(x_i), \beta) = \frac{exp(\beta_j^T H^*(x_i))}{\sum_{j=1}^{M} exp(\beta_j^T H^*(x_i))}$ for the ASMLBELM-WCFs and the ASMLNLELM-WCFs.

    (2) $P(y_{i,j} = 1 | K_{test}(x_i), \beta) = \frac{exp(\pi_j^T K_{test}(x_i))}{\sum_{k=1}^{M} exp(\pi_j^T K_{test}(x_i))}$ for the ASMLKELM-WCFs.

## D. The lower bound of the ASMLELM

In this section, the lower bound of the proposed ASMLELM will be derived. From (19), we have:

$$\nabla^2 L(\beta) \geq B \tag{41}$$

From (20), it is well known that $B$ is symmetric and negative definite independent from $\beta$. $\beta$ at the *(t+1)*-th iteration is defined as:

$$\beta^{t+1} = \hat{\beta}^t - (B - \lambda\Lambda^t)^{-1}\nabla L(\beta^t) \tag{42}$$

Equation (42) corresponds to the following equation:

$$Q(\beta) = (\beta - \beta')^T \nabla L(\beta') + \frac{1}{2}(\beta - \beta')^T B(\beta - \beta') - \lambda \parallel \beta \parallel_1 \tag{43}$$

From [29], (43) can be expressed as follows:

$$Q1(\beta) = (\beta - \beta')^T \nabla L(\beta') + \frac{1}{2}(\beta - \beta')^T (B - \lambda\Lambda)(\beta - \beta') \tag{44}$$

Then, we have the following two lemmas:

**Lemma 1**:

(a): $Q1(\beta)$ is maximized at: $\hat{\beta} = \beta' - (B - \lambda\Lambda^t)^{-1}\nabla L(\beta')$.

(b): $Q1(\hat{\beta}) = -\frac{1}{2}\nabla L^T(\beta')(B - \lambda\Lambda^t)^{-1}\nabla L(\beta') \geq 0$, where the inequality is strictly satisfied if $\nabla L(\beta') \neq 0$.

**Proof:**

(a) As $\nabla Q1(\beta) = \nabla L(\beta') + (B - \lambda\Lambda)(\beta - \beta') = 0$, we have $\hat{\beta} = \beta' - (B - \lambda\Lambda^t)^{-1}\nabla L(\beta')$.

(b) As $Q1(\hat{\beta}) = -((B - \lambda\Lambda^t)^{-1}\nabla L(\beta'))^T \nabla L(\beta') + \frac{1}{2}((B - \lambda\Lambda^t)^{-1}\nabla L(\beta'))^T (B - \lambda\Lambda^t)((B - \lambda\Lambda^t)^{-1}\nabla L(\beta'))$

$= -\nabla L(\beta')^T (B - \lambda\Lambda^t)^{-1}\nabla L(\beta') + \frac{1}{2}\nabla L(\beta')^T (B - \lambda\Lambda^t)^{-1}\nabla L(\beta') = -\frac{1}{2}\nabla L(\beta')^T (B - \lambda\Lambda^t)^{-1}\nabla L(\beta') \geq 0$,

the inequality is strictly satisfied if $\nabla L(\beta') \neq 0$.

**Lemma 2**:

(a) Monotonicity: $L(\beta^{t+1}) \geq L(\beta^t)$.

(b) Convergence: The sequence $\nabla L(\beta^t)$ converges to 0 if L is bounded as described in (a).

**Proof:**

(a) For the convenience, let $h = (B - \lambda \Lambda^t)^{-1} \nabla L(\beta^t)$. Then, we have

$$L(\beta^{t+1}) - L(\beta^t) = h^T \nabla L(\beta^t) + \frac{1}{2} h^T \nabla^2 L(\beta^t + \rho h) h \geq h^T \nabla L(\beta^t) + \frac{1}{2} h^T (B - \lambda \Lambda^t) h \geq 0.$$

(b) To prove this lemma, suppose that $\| \nabla L(\beta^t) \|$ is bounded by a value larger than 0. From (b) of Lemma 1, it can be seen that the increments are lower bounded. Therefore, it contradicts the boundedness of Q1. As a result, it can be concluded that the sequence $\nabla L(\beta^t)$ converges to 0.

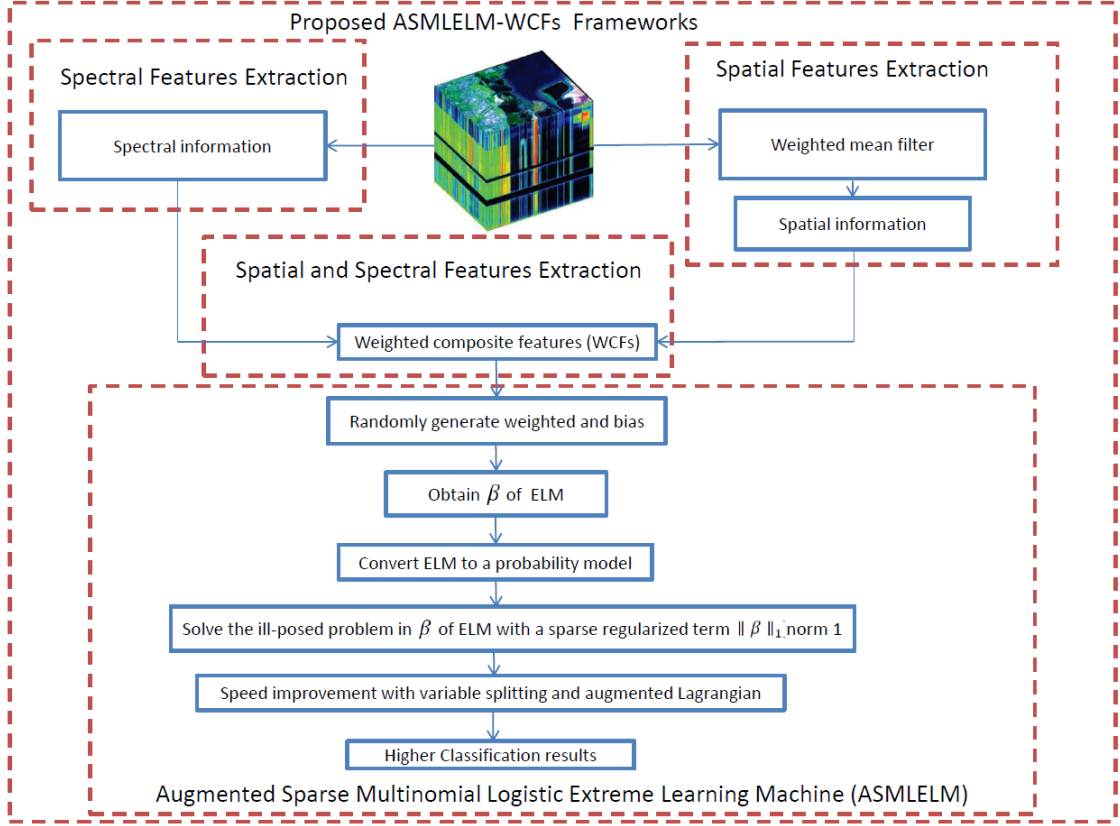

Figure 1. The flowchart of proposed ASMLELM-WCFs framework

## IV. Experiments and Analysis

### A. HSI Data Sets

In this section, the performances of the proposed new framework using two well-known HSI data sets (the Indian Pines data set and the Pavia University data set) will be evaluated. These two data sets are available in the public.

(1) Indian Pines data set: The Indian Pines HSI data set consists of the urban images collected by the AVIRIS sensors built in June 1992. The image scene has 145× 145 pixels with 200 spectral bands, where 20 spectral bands are the water absorption bands. Each band is ranging from 0.2μm to 2.4μm and the spatial resolution of the HSI data is 20m per pixel. There are totally 16 classes in the HSI data.

(2) Pavia University data set: The Pavia University HSI data set consists of data over the city Pavia Italy acquired by the ROSIS instrument in 2001. The image scene has 610×340 pixels with 103 spectral bands after removing 12 water absorption bands. The spatial resolution of the HSI data is 1.3m per pixel. There are totally 9 classes in the HSI data.

### B. Measurements and Parameter Setting

The parameter setting and the measurements are described below before conducting the experiments. The proposed six methods are compared with the state-of-the-art methods including SVM, SVM with CK (SVM-CK), LORSAL, KLORSAL, SMLR-SpATV, BELM, NLELM, and KELM. The LIBSVM [53] software is used for the implementation of the SVM and the SVM-CK. For the kernel based methods such as the SVM, the SVM-CK, the KELM, the ASMLKELM and the ASMLKELM-WCFs, the Gaussian kernel is used. The Gaussian kernel parameter $\sigma$ and the penalty parameter $C$ will be automatically tuned by the three folds cross validations. $C=2^p$, $\sigma = 2^q$, $p = \{1, 2,\dots, 11, 12, 13, 14, 15\}$ and $q = \{-6, -5, -4, -3, -2, -1, 0, 1\}$. Other parameters of the SVM and the SVM-CK are set the same as [1]. The parameters of the LORSAL, the KLORSAL (kernel based LORSAL) and the SMLR-SpATV (KLORSAL with the weighted Markov random field [54]-[55]) are chosen same as [43]. All experiments are conducted in MATLAB R2015a and run in a computer with 2.9 GHz CPU and 32.0 G RAM.

(1) For the proposed ASMLBELM, the total number of the neurons in the hidden layer $L$ and $\lambda$ are very important parameters. They will be evaluated in the next subsection.

(2) For the ASMLNLELM, the parameter C will be automatically tuned by these three folds cross validations. The effects of $L$ and $\lambda$ will be evaluated in the next subsection.

(3) The important parameters of the ASMLKELM are $C$, $\sigma$ and $\lambda$. $C$ and $\sigma$ will be automatically tuned by these three folds cross validations. The effect of $\lambda$ will be evaluated in the next subsection.

(4) $C$, $L$, $\lambda$ and μ are important parameters of the proposed ASMLBELM-WCFs. $C$ will be automatically tuned by these three folds cross validations. μ is empirically set to be 0.1 in the experiments. For the parameters $L$ and $\lambda$, their effects will be evaluated in the next subsection.

(5) The proposed ASMLNLELM-WCFs has several parameters such as $L$, $C$, $\lambda$ and μ. The parameter $C$ will be automatically tuned by these three folds cross validations. μ is empirically set to be 0.1 in the experiments. For the parameters $L$ and $\lambda$, their effects will be evaluated in the next subsection.

(6) For the ASMLKELM-WCFs, $C$ and $\sigma$ will be automatically tuned by these three folds cross validations. μ is empirically set to be 0.1 in the experiments. The effect of $\lambda$ will be evaluated in the next subsection.

*C. Discussion on Parameters*

In this subsection, several important parameters of the proposed methods will be evaluated and its performance will be compared with the BELM and the NLELM. It is worth noting that the window size is set at 9 for the WCFs based methods in both experiments 1 and experiment 2, which means the widths of the neighborhood window are set at 9. The effects of the window size for the proposed CF based methods will be evaluated after evaluating the effects of $\lambda$ and $L$.

**Experiment 1**: In Fig. 2, the effect of the parameter $\lambda$ ($\lambda = 2^a$) on the proposed method is evaluated. For convenience, in this experiments, the number of the hidden layer is set at $L$=550 and $L$=900 for the Indian Pines data set and the Pavia University data set, respectively. Fig. 1(a) and Fig. 1(b) plot the OA results as a function of a with 1043 and 3921 training samples (10% and 9% of the available samples) of the Indian Pines data set and the Pavia University data set, respectively. From Fig. 1, it can be seen that these three spectral and these three spatial classifiers are still stable to achieve the good performances when the parameter $a$ is varying. It can also be seen that the proposed spectral classifiers can achieve higher accuracy when $a$ is relatively large. Overall, the proposed six classifiers can achieve good and stable performances when the parameter $a$ is varying. For following experiments, both Indian Pines datasets and Pavia University datasets, we set $a$ to be -20 for the proposed ASMLBELM-WMFs, ASMLNLELM-WMFs and ASMLKELM-WMFs, and assign $a$ to be -10 for the proposed ASMLBELM and ASMLNLELM. For the proposed ASMLKELM, we set $a$ to be -17 in Indian Pines datasets, and set $a$ to be -13 in Pavia University datasets.

**Experiment 2**: In Fig. 3, the OA results are plotted as a function of the hidden layer neurons $L$ and its effects on the proposed methods, the BELM and the NLELM. From Fig. 2, it can be seen that the ASMLBELM and the ASMLNLELM always achieve higher accuracies than the BELM and the NLELM. It can also be seen that the proposed spectral spatial classifiers based on the ASMLBELM-WCFs and the ASMLNLELM-WCFs can greatly improve the performances of the proposed spectral classifiers based on the ASMLBELM and the ASMLNLELM. This is because the spatial information is considered in these two spectral classifiers. Therefore, the ASMLBELM, the ASMLNLELM, the ASMLBELM-WCFs and the ASMLNLELM-WCFs can achieve better performances compared with the BELM and the NLELM for any value of $L$. For all the next experiments, if no special emphasized, we set $L$ to be 450 for BELM, ASMLBELM-based (including ASMLBELM and ASMLBELM-WCFs), 1000 for NLELM, ASMLNLELM-based (including ASMLNLELM and ASMLNLELM-WCFs) in Indian Pines datasets. In Pavia University datasets, we set $L$ to be 1100 for BELM, NLELM, ASMLBELM, ASMLNLELM, ASMLBELM-WCFs and ASMLNLELM-WCFs.

**Experiment 3**: In this experiment, the effects of the widths of the window for the ASMBELM-WCFs, the ASMNLELM-WCFs and the ASMKELM-WCFs will be evaluated. Fig. 4 plots the OA result as a function of the window size of these three spectral spatial classifiers. From Fig. 3, it can be seen that the results of the proposed method are not very good when the window size is too small. However, the results of the proposed methods become very good and stable when the window size is large. This shows the generalization for achieving the good performance of the spectral spatial classifiers. For the convenience, the window size is set as 13 both for the Indian Pines data set and the Pavia University data set in the following experiments.

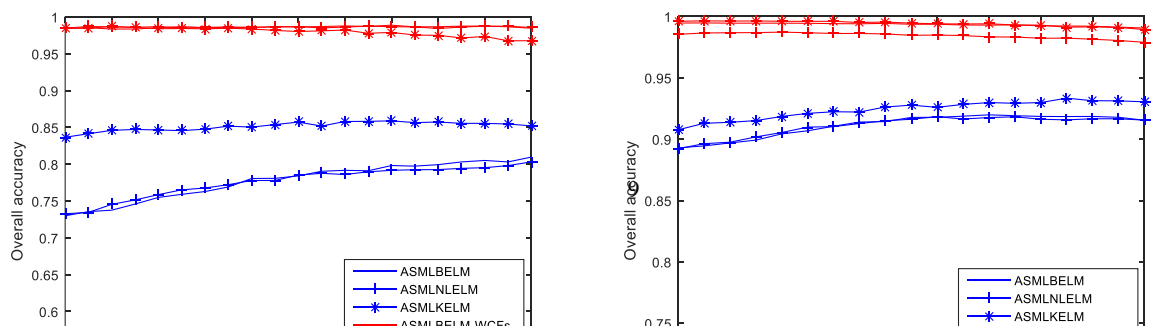

(a)            (b)

Fig. 2. Effect of the parameter *a* in an image in (a) the Indian Pines data set and (b) Pavia University data set.

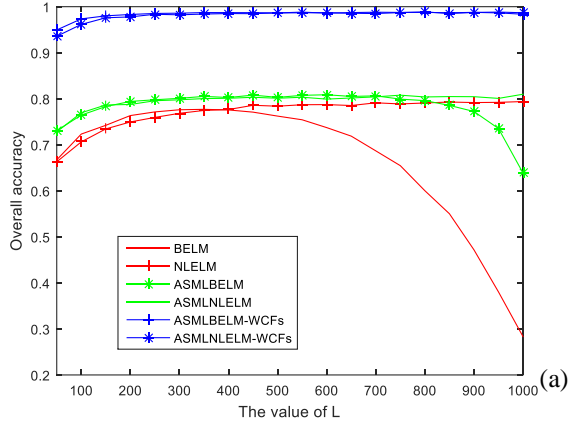 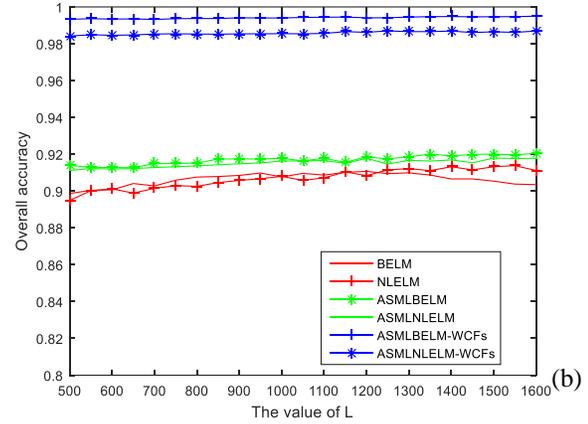

Fig. 3. Effect of the total number of neurons in the hidden layer L in an image in (a) the Indian Pines data set (b) the Pavia University data set.

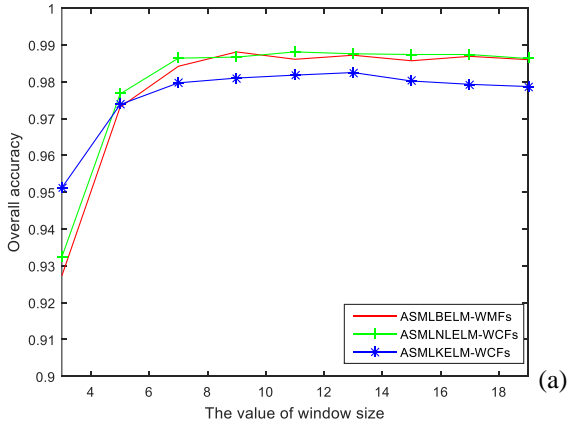 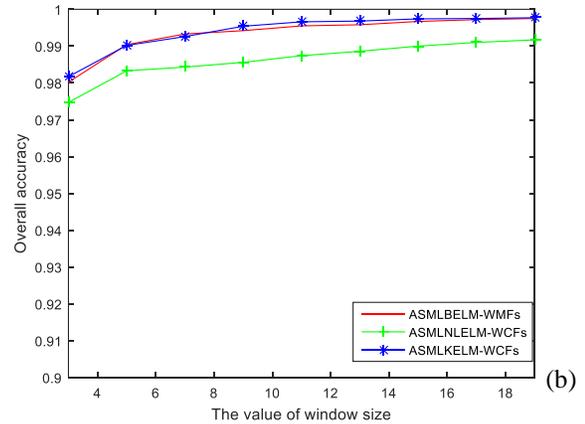

Fig. 4. Effect of the window size in an image in (a) the Indian Pines data set and (b) the Pavia University data set.

Table 1. Classification accuracy using 10% of the labeled samples per class for the Indian Pines data set (The best results are bolded).

| NO | Train | Test | SVM | SVM-CK | LORSAL | KLORSAL | SMLR-SpATV | BELM | NLELM | KELM | ASMLBELM | ASMLNLELM | ASMLKELM | ASMLBELM-WCFs | ASMLNLELM-WCFs | ASMLKELM-WCFs |
|---|---|---|---|---|---|---|---|---|---|---|---|---|---|---|---|---|
| 1 | 6 | 48 | 68.12 | 83.75 | 1.04 | 76.88 | 89.58 | 48.96 | 33.33 | 74.17 | 47.29 | 41.88 | 72.08 | **96.25** | 93.33 | 90.83 |
| 2 | 144 | 1290 | 83.17 | 95.48 | 73.36 | 82.25 | 95.95 | 76.55 | 81.09 | 83.27 | 81.03 | 80.81 | 83.29 | 98.40 | **98.62** | 97.86 |
| 3 | 84 | 750 | 75.52 | 95.77 | 49.44 | 68.76 | 98.01 | 56.80 | 59.44 | 70.31 | 63.07 | 63.44 | 74.44 | **98.52** | 98.19 | 97.89 |
| 4 | 24 | 210 | 72.19 | 93.1 | 22.00 | 60.19 | 98.95 | 44.76 | 43.10 | 66.9 | 48.29 | 45.67 | 63.71 | 98.09 | **99.05** | 96.57 |
| 5 | 50 | 447 | 92.73 | 95.41 | 84.88 | 89.66 | 95.95 | 86.80 | 87.99 | 92.15 | 88.01 | 88.17 | 92.48 | 97.70 | **98.32** | 97.58 |
| 6 | 75 | 672 | 96.06 | 99.24 | 94.43 | 95.33 | 99.08 | 94.87 | 96.89 | 96.49 | 96.13 | 96.19 | 95.95 | **99.63** | 99.58 | 99.16 |
| 7 | 3 | 23 | 75.22 | 77.39 | 0 | 35.22 | 52.61 | 4.78 | 1.30 | 71.74 | 7.39 | 9.57 | 70.00 | 90.00 | 88.69 | **93.04** |
| 8 | 49 | 440 | 98.73 | 97.5 | 98.77 | 98.59 | **100.0** | 99.07 | 99.52 | 98.84 | 99.57 | 99.45 | 98.93 | 99.75 | **99.93** | 99.72 |
| 9 | 2 | 18 | 67.78 | 77.78 | 0 | 46.67 | 5.00 | 3.89 | 3.89 | 58.33 | 0.56 | 1.67 | 49.44 | 91.67 | **96.11** | 87.22 |
| 10 | 97 | 871 | 77.89 | 93.64 | 55.76 | 71.55 | 94.08 | 64.88 | 67.83 | 80.11 | 65.48 | 66.79 | 79.39 | **97.63** | 97.50 | 95.91 |
| 11 | 247 | 2221 | 85.96 | 96.82 | 72.78 | 80.53 | 99.18 | 76.79 | 79.99 | 86.36 | 82.49 | 80.07 | 87.67 | 99.20 | **99.31** | 98.81 |
| 12 | 62 | 552 | 84.04 | 92.7 | 63.75 | 76.96 | 97.30 | 69.31 | 74.11 | 81.41 | 76.30 | 75.65 | 83.66 | 97.83 | **98.32** | 97.37 |
| 13 | 22 | 190 | 98.95 | 99.26 | 97.79 | 99.47 | 99.47 | 98.42 | 99.42 | 99.05 | 99.47 | 99.32 | 99.21 | 99.42 | 99.42 | **99.74** |

| NO. | Train | Test | SVM | SVM-CK | LORSAL | KLORSAL | SMLR-SpATV | BELM | NLELM | KELM | ASMLBELM | ASMLNLELM | ASMLKELM | ASMLBELM-WCFs | ASMLNLELM-WCFs | ASMLKELM-WCFs |
|---|---|---|---|---|---|---|---|---|---|---|---|---|---|---|---|---|
| 14 | 130 | 1164 | 95.82 | 98.51 | 94.60 | 95.84 | 99.23 | 94.28 | 96.20 | 96.34 | 95.52 | 95.49 | 96.31 | 99.80 | **99.91** | 99.73 |
| 15 | 38 | 342 | 61.02 | 93.36 | 64.01 | 68.10 | 98.16 | 61.26 | 65.35 | 60.94 | 65.58 | 66.46 | 65.47 | **99.27** | 99.15 | 98.92 |
| 16 | 10 | 85 | 93.29 | **96.82** | 55.06 | 74.94 | 86.47 | 45.30 | 71.06 | 79.88 | 74.94 | 76.35 | 76.24 | 91.53 | 94.35 | 89.18 |
| OA | | | 85.71 | 96.05 | 73.14 | 82.26 | 97.50 | 77.01 | 79.92 | 85.28 | 80.81 | 80.29 | 86.01 | 98.72 | **98.85** | 98.18 |
| AA | | | 82.91 | 92.91 | 57.98 | 76.31 | 88.11 | 64.17 | 66.28 | 81.02 | 68.19 | 67.94 | 80.52 | 97.17 | **97.49** | 96.22 |
| k | | | 83.69 | 95.50 | 69.15 | 79.74 | 97.15 | 73.62 | 76.97 | 83.18 | 77.98 | 77.42 | 84.00 | 98.55 | **98.69** | 97.92 |

Table 2. Classification accuracy using 9% of the labeled samples per class for the Pavia University data set (The best results are bolded).

| NO. | Train | Test | SVM | SVM-CK | LORSAL | KLORSAL | SMLR-SpATV | BELM | NLELM | KELM | ASMLBELM | ASMLNLELM | ASMLKELM | ASMLBELM-WCFs | ASMLNLELM-WCFs | ASMLKELM-WCFs |
|---|---|---|---|---|---|---|---|---|---|---|---|---|---|---|---|---|
| 1 | 548 | 6083 | 87.48 | 98.74 | 71.35 | 85.47 | 99.65 | 85.09 | 83.75 | 87.07 | 87.24 | 87.10 | 89.62 | 99.13 | 97.41 | 99.33 |
| 2 | 540 | 18109 | 88.95 | 99.11 | 76.44 | 88.64 | 98.72 | 92.53 | 92.08 | 94.01 | 93.29 | 92.86 | 94.25 | 99.84 | 99.59 | **99.88** |
| 3 | 392 | 1707 | 76.45 | 97.73 | 71.04 | 76.39 | 97.72 | 76.33 | 75.68 | 84.87 | 79.65 | 79.67 | 84.38 | **98.90** | 95.89 | 99.41 |
| 4 | 542 | 2540 | 97.09 | 99.24 | 95.72 | 96.93 | 97.69 | 96.46 | 97.22 | 97.87 | 97.58 | 97.47 | 97.66 | 99.32 | 99.19 | **99.62** |
| 5 | 265 | 1080 | 99.50 | **100.0** | 99.89 | 99.61 | 100.0 | 97.52 | 99.48 | 99.41 | 99.70 | 99.70 | 99.48 | 99.89 | 99.77 | 99.82 |
| 6 | 532 | 4497 | 88.75 | 99.55 | 77.21 | 87.15 | 99.99 | 92.70 | 95.17 | 93.94 | 93.90 | 94.53 | 99.99 | 99.75 | | **100.00** |
| 7 | 375 | 955 | 90.65 | 99.77 | 78.34 | 90.04 | 99.78 | 92.50 | 92.97 | 93.99 | 90.19 | 90.12 | 92.63 | **99.92** | 99.71 | 99.98 |
| 8 | 514 | 3168 | 88.14 | 97.26 | 75.35 | 82.16 | 99.25 | 89.48 | 90.78 | 89.77 | 86.49 | 87.49 | 88.59 | 98.90 | 96.64 | **99.18** |
| 9 | 231 | 716 | 99.58 | 98.58 | 90.66 | 88.74 | 92.11 | 99.68 | 99.76 | 99.83 | 99.82 | 99.73 | 99.92 | 99.76 | 99.89 | **99.82** |
| OA | | | 89.14 | 98.93 | 77.63 | 87.79 | 88.88 | 90.95 | 90.84 | 92.82 | 91.78 | 91.61 | 93.10 | 99.60 | 98.85 | **99.71** |
| AA | | | 90.73 | 98.89 | 81.78 | 88.35 | 98.32 | 91.36 | 91.72 | 93.55 | 91.99 | 92.01 | 93.45 | 99.52 | 98.65 | **99.67** |
| KAPPA | | | 85.44 | 98.54 | 70.73 | 83.77 | 98.48 | 87.80 | 87.68 | 90.29 | 88.88 | 88.68 | 90.66 | 99.45 | 98.43 | **99.60** |

### D. Experiment Results and Comparisons in the Indian Pines Data Set and the Pavia University Data Set

In this subsection, the classification results are evaluated using the Indian Pines data set and the Pavia University data set. Table 1 and Table 2 show the training samples and the testing samples of the Indian Pines data set and the Pavia University data set. 10% of the training samples are employed for the training and the remaining samples are employed for the testing in the Indian Pines data set. Similarly, 9% of the training samples are employed for the training and the remaining samples are employed for the testing in the Pavia University data set.

The classification accuracies are shown in Table 1 and Table 2. It can be seen that the ASMLBELM, the ASMLNLELM and ASMLKELM obtain the higher accuracies than the BELM the NLELM and the KELM, respectively. The performances of these proposed three spectral classifiers are improved dramatically when the spatial information (WCFs) are considered. Compared with other spectral spatial classifiers such as the SVM-CK and the SMLR-SpATV, the ASMLBELM-WCFs, the ASMLNLELM-WCFs and ASMLKELM-WCFs have achieved the higher classification accuracies than the SVM-CK and the SMLR-SpATV. Based on the above results, it can be concluded that the proposed six methods can achieve the very good performances. Fig. 5 and Fig. 6 show the images and the ground truth obtained by different methods for the Indian Pines data set and the Pavia University data set.

### E. Experiment Results with Different Number of Training Samples of The Proposed methods

The performances of these six methods are evaluated under different numbers of training samples. The total number of training samples is chosen as 5, 10, 15, 20, 25, 30, 35 and 40 from each class. Half of the total samples are chosen when the total number of training samples is more than half of the total samples of the class. For the parameters in the proposed methods, the approaches discussed in the previous subsections are employed.

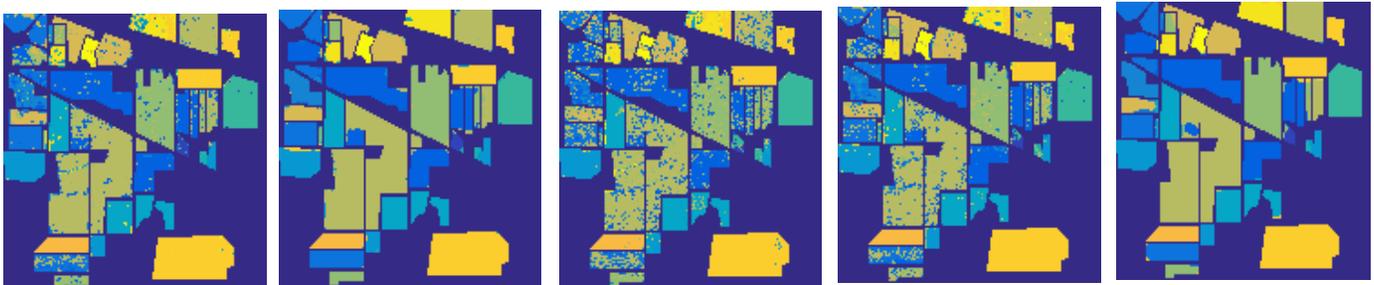

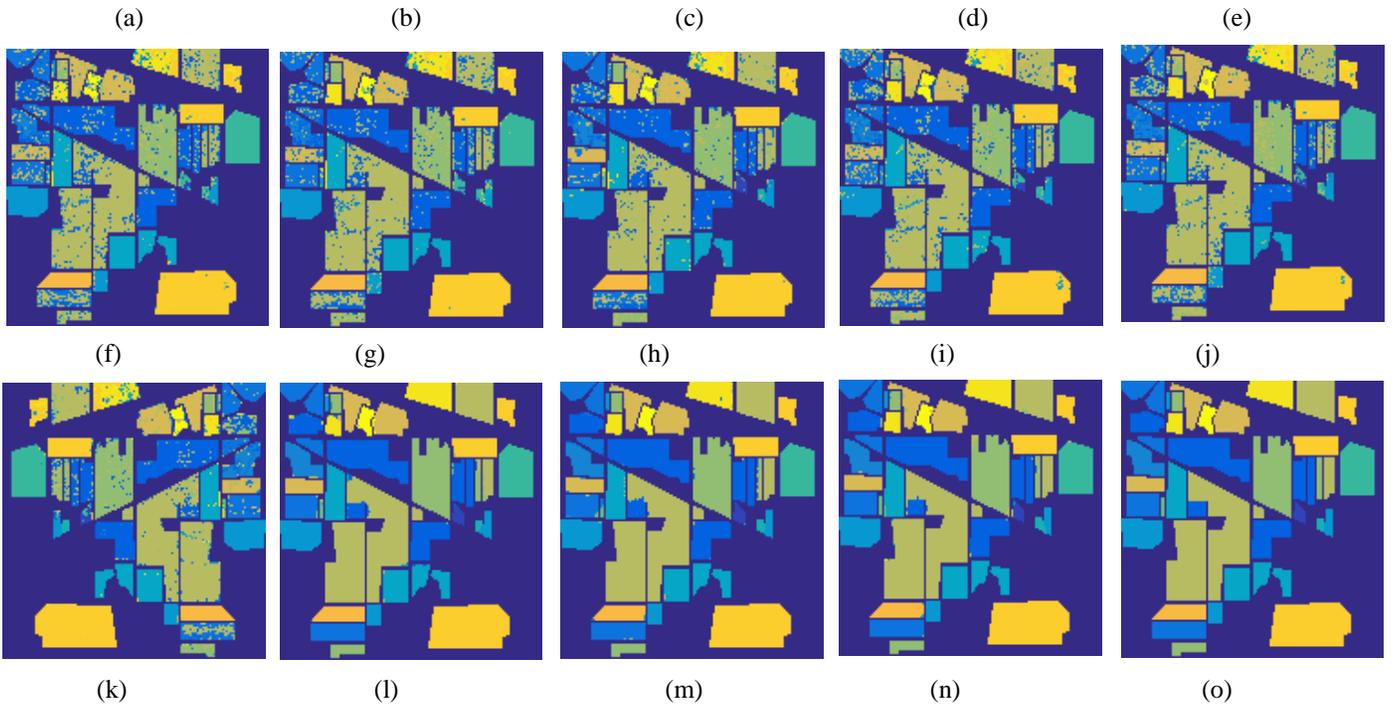

Fig. 5. Images of the Indian Pines data set. (a) The SVM (OA=85.71). (b) The SVM-CK (OA=96.05). (c) The LORSAL (OA=73.14). (d) The KLORSAL (OA=82.26). (e) The SMLR-SpATV (OA=97.50). (f) The BELM (OA=77.16). (g) The NLELM (OA=79.18). (h) The KELM (OA=85.28). (i) The ASMLBELM (OA=80.62). (j) The ASMLNLELM (OA=80.47). (k) The ASMLKELM (OA=85.93). (l) The ASMLBELM-WCFs (OA=98.33). (m) The ASMLNLELM-WCFs (OA=98.21). (n) The ASMLKELM-WCFs (OA=98.57) with 10% training samples. (o) The ground truth.

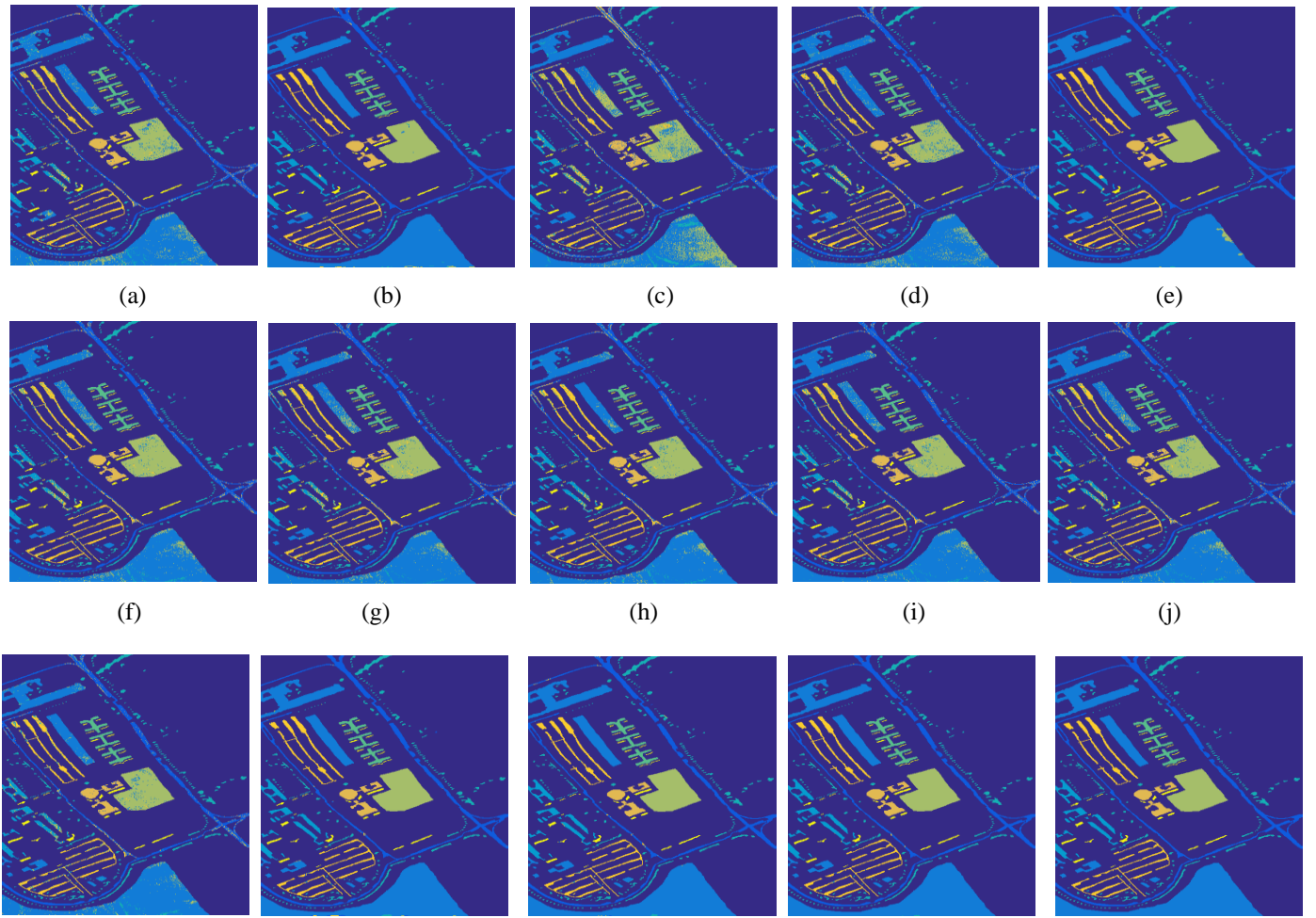

(k)  (l)  (m)  (n)  (o)

Fig. 6. Images of the Pavia University data set. (a) The SVM (OA=89.14). (b) The SVM-CK (OA=98.93). (c) The LORSAL (OA=77.63). (d) The KLORSAL (OA=87.79). (e) The SMLR-SpATV (OA=98.88). (f) The BELM (OA=90.87). (g) The NLELM (OA=90.80). (h) The KELM (OA=92.82). (i) The ASMLBELM (OA=92.06). (j) The ASMLNLELM (OA=91.71). (k) The ASMLKELM (OA=93.07). (l) The ASMLBELM-WCFs (OA=99.46). (m) The ASMLNLELM-WCFs (OA=99.44). (n) The ASMLKELM-WCFs (OA=99.69) with 9% training samples. (o) The ground truth.

From Table 3 and Table 4, it can be seen that the classification accuracies of these three spectral classifiers are better than those of the BELM, the NLELM and the KELM, respectively. These three spectral spatial classification algorithms can also achieve better performances. Compared with other classification methods listed in Table 3 and Table 4, these three spectral spatial classification methods achieve the higher accuracies than other spectral and spectral spatial classification methods. We can find an interesting result in Table 3 and Table 4 that the classification accuracies of BELM decrease when the number of training samples increases in most cases. This is because the ill-posed problem is particularly serious in BELM. Fortunately, the proposed ASMLBELM and ASMLNLELM-WCFs well alleviate this problem. In Table 3 and Table 4, we also display the execution time of the six proposed algorithms and other methods when using 100 training samples. From Table 3 and Table 4, in Indian Pines datasets and Pavia University datasets, the three proposed spectral algorithms, i.e. ASMLBELM, ASMLBELM and ASMLKELM need more consuming time than BELM, NLELM and KELM, respectively. But these three proposed algorithms have less consuming time than SVM. In Indian Pines data set, the proposed ASMLEBELM-WCFs and ASMLNLELM-WCFs algorithms need less consuming time than SVM-CK and SMLR-SpATV. Although the proposed ASMLEKELM-WCFs needs more consuming time than SMLR-SpATV, it needs less consuming time than SVM-CK. In Pavia University datasets, the three proposed spectral spatial algorithms have less consuming time than SVM-CK and SMLR-SpATV. Based on the above analysis, it can be seen that the proposed six classification methods achieve the very good performances especially the three spectral spatial classification algorithms.

Table 3. Classification accuracy (%) under different numbers of training samples for the Indian Pines data set (The best results are bolded).

| Number | Index | SVM | SVM-CK | LORSAL | KLORSAL | SMLR-SpATV | BELM | NLELM | KELM | ASMLBELM | ASMLNLELM | ASMLKELM | ASMLBELM-WCFs | ASMLNLELM-WCFs | ASMLKELM-WCFs |
|---|---|---|---|---|---|---|---|---|---|---|---|---|---|---|---|
| 5 | OA | 53.84 | 57.30 | 43.53 | 57.34 | 69.67 | 42.22 | 50.94 | 54.68 | 53.60 | 54.86 | 54.95 | 77.96 | **79.82** | 72.94 |
| | AA | 67.36 | 70.78 | 53.28 | 68.89 | 81.98 | 54.75 | 65.52 | 66.6 | 67.73 | 68.23 | 67.61 | 87.09 | **87.66** | 81.85 |
| | k | 48.54 | 52.52 | 37.84 | 52.59 | 66.16 | 36.19 | 45.62 | 49.4 | 48.33 | 49.57 | 49.69 | 75.28 | **77.30** | 69.55 |
| 10 | OA | 62.85 | 69.94 | 49.05 | 65.48 | 78.60 | 43.04 | 58.80 | 63.45 | 60.24 | 61.39 | 63.60 | 86.78 | **86.22** | 81.73 |
| | AA | 74.01 | 80.45 | 62.77 | 75.99 | 88.22 | 55.12 | 72.85 | 75.63 | 74.04 | 74.20 | 75.38 | 92.76 | **92.52** | 89.13 |
| | k | 58.37 | 66.30 | 43.7 | 61.38 | 76.03 | 36.92 | 54.21 | 59.11 | 55.57 | 56.72 | 59.32 | 85.08 | **84.46** | 79.40 |
| 15 | OA | 69.77 | 78.15 | 55.3 | 68.04 | 83.59 | 42.39 | 61.64 | 67.87 | 62.76 | 64.67 | 68.86 | 89.81 | **90.06** | 87.04 |
| | AA | 80.10 | 86.66 | 67.06 | 78.49 | 91.02 | 52.42 | 75.13 | 79.66 | 76.01 | 77.83 | 79.71 | 94.62 | **94.71** | 93.10 |
| | k | 66.07 | 75.38 | 50.18 | 64.23 | 81.48 | 35.72 | 57.27 | 63.99 | 58.27 | 60.42 | 65.09 | 88.46 | **88.74** | 85.34 |
| 20 | OA | 72.54 | 80.86 | 58.83 | 72.27 | 86.73 | 38.71 | 65.24 | 71.34 | 64.05 | 67.25 | 71.62 | 92.00 | **92.30** | 88.93 |
| | AA | 81.77 | 88.35 | 69.62 | 81.62 | 93.10 | 45.46 | 77.86 | 81.67 | 77.09 | 79.55 | 81.57 | 95.77 | **96.03** | 94.30 |
| | k | 69.06 | 78.41 | 53.96 | 68.85 | 85.00 | 31.54 | 61.06 | 67.79 | 59.71 | 63.22 | 68.06 | 90.90 | **91.25** | 87.47 |
| 25 | OA | 74.89 | 83.52 | 60.16 | 72.92 | 87.82 | 30.80 | 66.57 | 72.04 | 62.07 | 68.93 | 74.21 | 92.77 | **93.28** | 91.82 |
| | AA | 83.64 | 90.12 | 70.98 | 82.99 | 93.81 | 34.88 | 79.60 | 82.86 | 74.54 | 81.04 | 83.78 | 96.24 | **96.65** | 95.35 |
| | k | 71.74 | 81.36 | 55.4 | 69.58 | 86.20 | 22.64 | 62.59 | 68.59 | 57.35 | 65.09 | 70.97 | 91.78 | **92.36** | 90.70 |
| 30 | OA | 75.10 | 85.31 | 61.89 | 73.93 | 89.71 | 16.60 | 67.15 | 75.24 | 44.70 | 69.87 | 76.52 | **94.52** | 94.12 | 92.47 |
| | AA | 84.16 | 91.88 | 71.16 | 82.95 | 94.48 | 14.68 | 80.20 | 84.5 | 52.11 | 81.52 | 85.52 | **97.14** | 96.96 | 96.15 |
| | k | 71.97 | 83.35 | 57.35 | 70.65 | 87.85 | 7.37 | 63.29 | 72.05 | 37.77 | 66.10 | 73.51 | **93.76** | 93.31 | 91.43 |
| 35 | OA | 76.74 | 87.22 | 63.7 | 75.09 | 89.67 | 40.01 | 68.75 | 75.73 | 66.67 | 71.41 | 76.99 | 95.41 | **95.53** | 93.30 |
| | AA | 85.15 | 92.89 | 72.89 | 84.05 | 95.06 | 46.22 | 80.65 | 85.55 | 78.74 | 82.88 | 85.59 | 97.46 | **97.64** | 96.37 |
| | k | 73.74 | 85.49 | 59.28 | 71.92 | 88.28 | 32.63 | 65.05 | 72.64 | 62.49 | 67.82 | 74.00 | 94.75 | **94.89** | 92.36 |
| 40 | OA | 78.47 | 88.9 | 64.25 | 75.93 | 90.74 | 53.93 | 70.19 | 76.58 | 71.33 | 72.30 | 77.67 | 95.70 | **96.09** | 94.57 |
| | AA | 86.01 | 93.75 | 73.76 | 83.57 | 95.20 | 64.58 | 81.31 | 86.09 | 82.30 | 82.62 | 85.96 | 97.56 | **97.84** | 96.97 |
| | k | 75.64 | 87.38 | 59.91 | 72.77 | 89.45 | 48.34 | 66.61 | 73.55 | 67.67 | 68.77 | 74.74 | 95.08 | **95.53** | 93.79 |
| 100 | Time(s) | 141.0 | 196. | 0.5 | 0.8 | 34.9 | **0.4** | 2.28 | 10.3 | 1.2 | 4.0 | 13.5 | 12.0 | 15.7 | 81.7 |

Table 4. Classification accuracy (%)under different numbers of training samples for the Pavia University data set (Best results are bolded).

| Number | Index | SVM | SVM-CK | LORSAL | KLORSAL | SMLR-SpATV | BELM | NLELM | KELM | ASMLBELM | ASMNLELM | ASMLKELM | ASMLBELM-WCFs | ASMNLELM-WCFs | ASMLKELM-WCFs |
|---|---|---|---|---|---|---|---|---|---|---|---|---|---|---|---|
| 5 | OA | 56.75 | 63.85 | 47.58 | 56.5 | 65.87 | 57.75 | 67.94 | 62.86 | 61.22 | 65.88 | 63.61 | **75.68** | 71.48 | 70.52 |
|  | AA | 69.79 | 72.74 | 47.3 | 67.48 | 74.82 | 67.36 | 72.18 | 72.87 | 71.63 | 73.39 | 72.80 | **80.95** | 76.81 | 78.19 |
|  | k | 47.58 | 55.26 | 35.93 | 55.53 | 57.58 | 48.56 | 58.64 | 53.94 | 52.58 | 57.57 | 54.48 | **69.39** | 64.19 | 62.98 |
| 10 | OA | 66.31 | 73.82 | 50.19 | 62.49 | 76.23 | 57.75 | 72.53 | 69.21 | 72.67 | 70.92 | 71.01 | **83.51** | 76.43 | 81.30 |
|  | AA | 75.47 | 80.64 | 50.09 | 71.65 | 79.95 | 64.96 | 77.66 | 78.98 | 78.05 | 78.08 | 79.17 | **86.60** | 81.53 | 86.69 |
|  | k | 57.79 | 67.06 | 38.68 | 53.81 | 69.70 | 48.08 | 65.08 | 61.68 | 65.33 | 63.40 | 63.64 | **78.79** | 70.08 | 76.04 |
| 15 | OA | 70.58 | 80.18 | 55.61 | 66.11 | 80.96 | 59.98 | 73.74 | 72.77 | 73.59 | 74.26 | 72.71 | **86.77** | 83.15 | 85.84 |
|  | AA | 78.42 | 84.65 | 53.82 | 74.39 | 87.57 | 63.74 | 79.77 | 82.13 | 79.29 | 79.77 | 80.63 | **89.67** | 85.83 | 90.23 |
|  | k | 62.98 | 74.69 | 44.53 | 53.93 | 75.82 | 50.12 | 66.72 | 66.05 | 66.52 | 67.30 | 65.53 | **82.98** | 78.25 | 81.89 |
| 20 | OA | 72.45 | 84.2 | 59.27 | 70.6 | 85.20 | 61.94 | 75.04 | 76.75 | 73.91 | 76.76 | 76.95 | **91.00** | 85.14 | 90.40 |
|  | AA | 79.38 | 87.53 | 56.6 | 76.41 | 89.22 | 64.67 | 80.79 | 83.38 | 80.15 | 81.51 | 83.59 | **91.86** | 87.75 | 92.67 |
|  | k | 65.07 | 79.63 | 48.3 | 59.52 | 81.00 | 52.36 | 68.35 | 70.48 | 67.04 | 70.34 | 70.79 | **88.24** | .80.80 | 87.52 |
| 25 | OA | 73.90 | 89.05 | 58.33 | 71.44 | 88.48 | 59.67 | 77.56 | 78.95 | 77.56 | 78.09 | 79.02 | 92.33 | 88.17 | **92.53** |
|  | AA | 81.35 | 90.31 | 57.49 | 78 | 91.34 | 61.82 | 82.39 | 84.92 | 81.51 | 82.09 | 84.56 | 93.42 | 89.76 | **93.68** |
|  | k | 67.04 | 85.71 | 47.56 | 61.02 | 85.14 | 50.07 | 71.33 | 73.15 | 71.24 | 71.92 | 73.21 | 89.98 | 84.60 | **90.19** |
| 30 | OA | 77.70 | 89.21 | 60.69 | 73.64 | 91.02 | 61.35 | 78.87 | 78.53 | 78.36 | 79.90 | 80.37 | **93.31** | 90.23 | 93.03 |
|  | AA | 82.45 | 90.33 | 60.16 | 78.87 | 92.55 | 61.53 | 83.58 | 84.91 | 82.44 | 83.60 | 85.99 | **93.95** | 91.28 | 94.27 |
|  | k | 71.38 | 85.87 | 50.47 | 68.29 | 88.26 | 51.76 | 72.95 | 72.72 | 72.26 | 74.15 | 74.94 | **91.24** | 87.24 | 90.86 |
| 35 | OA | 75.19 | 90.42 | 61.71 | 75.16 | 91.55 | 59.01 | 79.66 | 81.09 | 78.44 | 79.55 | 81.32 | **94.67** | 91.13 | 94.19 |
|  | AA | 82.99 | 91.8 | 61.86 | 79.33 | 93.36 | 59.09 | 84.58 | 86.33 | 82.79 | 83.46 | 86.71 | **94.95** | 92.33 | 95.15 |
|  | k | 68.69 | 87.48 | 51.68 | 67.11 | 88.97 | 49.28 | 73.97 | 75.78 | 72.49 | 73.73 | 76.11 | **92.98** | 88.41 | 92.36 |
| 40 | OA | 77.09 | 91.9 | 63.15 | 76.92 | 91.68 | 56.78 | 80.21 | 82.53 | 79.77 | 80.91 | 82.65 | **95.64** | 92.41 | 95.54 |
|  | AA | 83.02 | 92.47 | 63.3 | 80.81 | 92.40 | 55.91 | 84.12 | 87.32 | 83.27 | 84.97 | 87.59 | **95.41** | 92.91 | 96.02 |
|  | k | 70.73 | 89.36 | 53.3 | 70.85 | 89.13 | 46.73 | 74.48 | 77.53 | 74.01 | 75.50 | 77.75 | **94.23** | 90.01 | 94.11 |
| 100 | Time(s) | 42.1 | 81.81 | 0.4 | 1.5 | 99.4 | 1.8 | 2.3 | 5.3 | 3.4 | 3.4 | 6.8 | 34.5 | 36.3 | 51.0 |

## V. Conclusion

This paper proposes three spectral algorithms and three spectral spatial methods which are improvements of ELM for performing the HSI classification. First, the ELM is represented by a probabilistic model via the MAP. Then, it was represented by a concave logarithmic likelihood function where its maximum can be obtained. Second, the Laplacian prior is adopted to it. The performance of the proposed framework is improved. Third, we adopt the LORSAL algorithm for the proposed framework in order to improve the performances. Fourthly, the spatial information are considered and the spectral spatial framework is employed for performing the HSI classification to further improve the classification accuracy. Finally, the lower bounds of the proposed method is derived by a rigorous mathematical proof, which demonstrate the good performances of proposed six proposed algorithms.

For future work, we will focus on improving the computationally of proposed ASMLKELM by further spare representation. Moreover, we will also develop the classification accuracy by resorting to extended multi-attribute profiles [56] (EMAPs) method.

## ACKNOWLEDGMENT

This work is supported by the National Nature Science Foundation of China (no. 61471132, 61372173), the Training program for outstanding young teachers in higher education institutions of Guangdong Province (no. YQ2015057).

## REFERENCES

[1] Y. Zhou, J. Peng, C. L. P. Chen, "Extreme learning machine with composite kernels for hyperspectral image classification," IEEE Journal of Selected Topics in Applied Earth Observations and Remote Sensing, vol. 8, no. 6, pp. 2351-2360, 2015.

[2] A. Plaza, J. A. Benediktsson, J. W. Boardman, et al, "Recent advances in techniques for hyperspectral image processing," Remote sensing of environment, vol. 113, pp. S110-S122, 2009.

[3] Hughes G, "On the mean accuracy of statistical pattern recognizers," IEEE transactions on information theory, vol. 14, no. 1, pp. 55-63, 1968.

[4] J. Li, P. R. Marpu, A. Plaza, et al, "Generalized composite kernel frame work for hyperspectral image classification," IEEE Transactions on Geoscience and Remote Sensing, vol. 51, no. 9, pp. 4816-4829, 2013.

[5] T. Qiao, J. Ren et al, "Effective denoising and classification of hyperspectral images using curvelet transform and singular spectrum analysis," IEEE Trans. Geoscience and Remote Sensing, vol. 55, no. 1, pp. 119-133, 2017.

[6] J. Zabalza, J. Ren, J. Zheng, J. Han, H. Zhao, S. Li, and S. Marshall, "Novel two dimensional singular spectrum analysis for effective feature extraction and data classification in hyperspectral imaging," IEEE Trans. Geoscience and Remote Sensing, vol. 53, no. 8, pp. 4418-4433, 2015.

[7] T. Qiao, J. Ren, C. Craigie, Z. Zabalza, C. Maltin, S. Marshall, "Singular spectrum analysis for improving hyperspectral imaging based beef eating quality evaluation," Computers and Electronics in Agriculture, 2015.

[8] J. Zabalza, J. Ren, Z. Wang, H. Zhao, J. Wang, and S. Marshall, "Fast implementation of singular spectrum analysis for effective feature extraction in hyperspectral imaging," IEEE Journal of Selected Topics in Earth Observation and Remote Sensing, vol. 8, no. 6, pp. 2845-53, 2015.

[9] F. Melgani, L. Bruzzone, "Classification of hyperspectral remote sensing images with support vector machines," IEEE Transactions on geoscience and remote sensing, vol. 42, no. 8, pp. 1778-1790, 2004.

[10] Q. He, C. Du, Q. Wang, F. Zhuang, and Z. Shi, "A parallel incremental extreme SVM classifier," Neurocomputing, vol. 74, pp. 2532–2540, 2011.

[11] J. Zabalza, J. Ren, J. Zheng, H. Zhao, C. Qing, Z. Yang, P. Du and S. Marshall, "Novel segmented stacked autoencoder for effective dimensionality reduction and feature extraction in hyperspectral imaging," Neurocomputing, vol. 185, pp. 1-10, 2016

[12] J. Ren, Z. Zabalza, S. Marshall and J. Zheng, "Effective feature extraction and data reduction with hyperspectral imaging in remote sensing," IEEE Signal Processing Magazine, vol. 31, no. 4, pp. 149-154, 2014.

[13] J. Zabalza, J.-C. Ren, J. Ren, Z. Liu, and S. Marshall, "Structured cova iance principle component analysis for real-time onsite feature extraction and dimensionality reduction in hyperspectral imaging," Applied Optics, vol. 53, no. 20, pp. 4440-4449, 2014.

[14] J. Zabalza, J. Ren, M. Yang, Y. Zhang, J. Wang, S. Marshall, J. Han, "Novel Folded-PCA for Improved Feature Extraction and Data Reduction with Hyperspectral Imaging and SAR in Remote Sensing," ISPRS J. Photogrammetry & Remote Sensing, vol. 93, no. 7, pp. 112-122, 2014.

[15] L. Fang, S. Li, W. Duan, J. Ren, J. Atli Benediktsson, "Classification of hyperspectral images by exploiting spectral-spatial information of superpixel via multiple kernels," IEEE Trans. Geoscience and Remote Sensing, vol. 53, no. 12, pp. 6663-6674, 2015.

[16] J. Li, JM. Bioucas-Dias, A. Plaza, "Spectral–spatial classification of hyperspectral data using loopy belief propagation and active learning," IEEE Transactions on Geoscience and Remote Sensing, vol. 51, no. 2, pp. 844-856, 2013.

[17] G. B. Huang, Q. Y. Zhu, C. K. Siew, "Extreme learning machine: theory and applications," Neurocomputing, vol. 70, no. 1, pp. 489-501, 2006.

[18] G. B. Huang, "An insight into extreme learning machines: random neurons, random features and kernels," Cognitive Computation, vol. 6, no. 3, pp. 376-390, 2014.

[19] G. B. Huang, H. Zhou, X. Ding, et al, "Extreme learning machine for regression and multiclass classification," IEEE Transactions on Systems, Man, and Cybernetics, Part B (Cybernetics), vol. 42, no. 2, pp. 513-529, 2012.

[20] Y. Wang, F. Cao, and Y. Yuan, "A study on effectiveness of extreme learning machine," Neurocomputing, vol. 74, pp. 2483–2490, 2011.

[21] H. J. Rong, Y. S. Ong, A. H. Tan, and Z. Zhu, "A fast pruned-extreme learning machine for classification problem," Neurocomputing, vol. 72, pp. 359–366, 2008.

[22] G. B. Huang, X. Ding, and H. Zhou, "Optimization method based extreme learning machine for classification," Neurocomputing, vol. 74, pp. 155–163, 2010.

[23] A. Samat, P. Du, S. Liu, et al, "$E^2$LMs: Ensemble Extreme Learning Machines for Hyperspectral Image Classification," IEEE Journal of Selected Topics in Applied Earth Observations and Remote Sensing, vol. 7, no. 4, pp. 1060-1069, 2014.

[24] Shen Y, Xu J, Li H, et al. ELM-based spectral-spatial classification of hyperspectral images using bilateral filtering information on spectral band-sub-sets[C]//Geoscience and Remote Sensing Symposium (IGARSS), 2016 IEEE International. IEEE, 2016: 497-500.

[25] Argüello F, Heras D B. ELM-based spectral–spatial classification of hyperspectral images using extended morphological profiles and composite feature mappings[J]. International Journal of Remote Sensing, 2015, 36(2): 645-664.

[26] Duan W, Li S, Fang L. Spectral-spatial hyperspectral image classification using superpixel and extreme learning machines[C]//Chinese Conference on Pattern Recognition. Springer, Berlin, Heidelberg, 2014: 159-167.

[27] Heras D B, Argüello F, Quesada-Barriuso P. Exploring ELM-based spatial–spectral classification of hyperspectral images[J]. International Journal of Remote Sensing, 2014, 35(2): 401-423.

[28] Chen C, Li W, Su H, et al. Spectral-spatial classification of hyperspectral image based on kernel extreme learning machine[J]. Remote Sensing, 2014, 6(6): 5795-5814.

[29] B. Krishnapuram, L. Carin, M. A. T. Figueiredo, et al, "Sparse multinomial logistic regression: Fast algorithms and generalization bounds," IEEE transactions on pattern analysis and machine intelligence, vol. 27, no. 6, pp. 957-968, 2005.

[30] D. L. Donoho, M. Elad, "Optimally sparse representation in general (nonorthogonal) dictionaries via ℓ1 minimization," Proceedings of the National Academy of Sciences, vol. 100, no. 5, pp. 2197-2202, 2003.

[31] J. Friedman, T. Hastie, S. Rosset, et al, "Discussion of boosting papers," Ann. Statist, vol. 32, pp. 102-107, 2004.

[32] A. Y. Ng, "Feature selection, L 1 vs. L 2 regularization, and rotational invariance," Proceedings of the twenty-first international conference on Machine learning. ACM, vol. 78, 2004.

[33] Afonso M V, Bioucas-Dias J M, Figueiredo M A T. Fast image recovery using variable splitting and constrained optimization[J]. IEEE Transactions on Image Processing, 2010, 19(9): 2345-2356.

[34] G. Camps-Valls, L. Gomez-Chova, J. Muñoz-Mari, et al, "Composite kernels for hyperspectral image classification," IEEE Geoscience and Remote Sensing Letters, vol. 3, no. 1, pp. 93-97, 2006.

[35] Y. Zhou, J. Peng, C. L. P. Chen, "Dimension reduction using   spatial and spectral regularized local discriminant embedding for hyperspectral image classification," IEEE Transactions on Geoscience and Remote Sensing, vol. 53, no. 2, pp. 1082-1095, 2015.

[36] R. Meir, T. Zhang, "Generalization error bounds for Bayesian mixture algorithms,"Journal of Machine Learning Research, vol. 4, pp. 839-860, 2003.

[37] M. Seeger, "Pac-bayesian generalisation error bounds for gaussian process classification," Journal of machine learning research, vol. 3, pp. 233-269, 2002.

[38] P. L. Bartlett, "The sample complexity of pattern classification with neural networks: the size of the weights is more important than the size of the network," IEEE transactions on Information Theory, vol. 44, no. 2, pp. 525-536,1998.

[39] C. Chen. A rapid supervised learning neural network for function interpolation and approximation. IEEE Transactions on Neural Networks, 1996, 7(5): 1220-1230.

[40] D. Bohning, "Multinomial logistic regression algorithm," An nals of the Institute of Statistical Mathematics, vol. 44, no. 1, pp. 197-200, 1992.

[41] D. Bohning, B. G. Lindsay, "Monotonicity of quadratic-approximation algorithms," Annals of the Institute of Statistical Mathematics, vol. 40, no. 4, pp. 641-663, 1988.

[42] T. P. Minka, "A comparison of numerical optimizers for logistic regression," Unpublished draft, 2003.

[43] D. R. Hunter and K. Lange, "A tutorial on MM algorithms," Amer. Statistician, vol. 58, no. 1, pp. 30-37, Feb, 2004.

[44] J. Li, J. M. Bioucas-Dias, A. Plaza, "Semisupervised hyperspectral image segmentation using multinomial logistic regression with active learning," IEEE Transactions on Geoscience and Remote Sensing, vol. 48, no. 11, pp. 4085-4098, 2010.

[45] J. Li, J. M. Bioucas-Dias, A. Plaza, "Hyperspectral image segmentation using a new Bayesian approach with active learning," IEEE Transactions on Geoscience and Remote Sensing, vol. 49, no. 10, pp. 3947-3960, 2011.

[46] J. Li, J. M. Bioucas-Dias, A. Plaza, "Semi-supervised hyperspectral image classification based on a Markov random field and sparse multinomial logistic regression," Geoscience and Remote Sensing Symposium, 2009 IEEE International, IGARSS 2009. IEEE, 3: III-817-III-820, 2009.

[47] B. Krishnapuram, D. Williams, Y. Xue, et al, "On Semi-Supervised Classification," NIPS, vol. 17, pp. 721-728. 2004

[48] J. Bioucas-Dias, M. Figueiredo, "Logistic regression via variable splitting and augmented lagrangian tools," Instituto Superior Técnico, TULisbon, Tech. Rep, 2009.

[49] L. Sun, Z. Wu,  J. Liu, et al, "Supervised spectral–spatial hyperspectral image classification with weighted Markov random fields," IEEE Transactions on Geoscience and Remote Sensing, vol. 53, no. 3, pp. 1490-1503, 2015.

[50]Bioucas-Dias J M, Figueiredo M A T. Multiplicative noise removal using variable splitting and constrained optimization[J]. IEEE Transactions on Image Processing, 2010, 19(7): 1720-1730.

[51] J. Eckstein, D. P. Bertsekas, "On the Douglas—Rachford splitting method and the proximal point algorithm for maximal monotone operators," Mathematical Programming, vol. 55, no. 1, pp. 293-318, 1992,

[52] M. V. Afonso, J. M. Bioucas-Dias, M. A.T. Figueiredo, "Fast image recovery using variable splitting and constrained optimization," IEEE Transactions on Image Processing, vol. 19, no. 9, pp. 2345-2356, 2010.

[53] C. C. Chang, C. J. Lin, "LIBSVM: a library for support vector machines," ACM Transactions on Intelligent Systems and Technology (TIST), vol. 2, no. 3, pp. 27, 2011.

[54] J. S. Yedidia, W. T. Freeman, Y. Weiss, "Understanding belief propagation and its generalizations," Exploring artificial intelligence in the new millennium, vol. 8, pp. 236-239, 2003.

[55] J. S. Yedidia, W. T. Freeman, Y. Weiss, "Constructing free-energy approximations and generalized belief propagation algorithms," IEEE Transactions on Information Theory, vol. 51, no. 7, pp. 2282-2312, 2005.

[56] Mura. M. Dalla, J. A. Benediktsson, B. Waske, et al, "Morphological attribute profiles for the analysis of very high resolution images," IEEE Transactions on Geoscience and Remote Sensing, vol. 48, no. 10, pp. 3747-3762, 2010.